\documentclass[11pt]{article}
\usepackage[utf8]{inputenc}
\usepackage[margin=1in]{geometry}

\usepackage{microtype}
\usepackage{graphicx}
\usepackage{subfigure}
\usepackage{booktabs} 
\usepackage{natbib}
\setcitestyle{open={(},close={)}}

\usepackage[colorlinks=true,citecolor=blue]{hyperref}

\usepackage{graphicx}
\usepackage{booktabs}
\usepackage{tikz}
\usepackage{comment}
\usepackage{amsmath,amssymb} 
\usepackage{color}
\usepackage{multirow}
\usepackage{subfigure}
\usepackage[accsupp]{axessibility}  

\begin{document}

\title{\bf Degenerate Swin to Win: Plain Window-based Transformer without Sophisticated Operations}

\author{\vspace{0.5in}\\\bf Tan Yu, Ping Li\\\\
Cognitive Computing Lab\\
Baidu Research\\
10900 NE 8th St. Bellevue, WA 98004, USA\\\\
\{tan.yu1503, pingli98\}@gmail.com 
}

\date{}

\maketitle

\begin{abstract}\vspace{0.3in}
\noindent The formidable accomplishment of Transformers in natural language processing has motivated the researchers in the computer vision community to build Vision Transformers. Compared with the Convolution Neural Networks (CNN), a Vision Transformer has a larger receptive field which is capable of characterizing the long-range dependencies. Nevertheless, the large receptive field of Vision Transformer is accompanied by the huge computational cost. To boost efficiency, the window-based Vision Transformers emerge. They crop an image into several local windows, and the self-attention is conducted within each window. To bring back the  global receptive field,  window-based Vision Transformers have devoted  a lot of efforts to achieving  cross-window communications by developing several sophisticated operations.  In this work, we check the necessity of  the key design element of Swin Transformer, the shifted window partitioning. We discover that a simple depthwise convolution is sufficient for achieving effective cross-window communications. Specifically, with the existence of the depthwise convolution, the shifted window configuration in Swin Transformer cannot lead to an additional performance improvement. Thus, we degenerate the Swin Transformer to a plain Window-based (Win) Transformer by discarding sophisticated shifted window partitioning. The proposed Win Transformer is  conceptually simpler and easier for implementation than Swin  Transformer. Meanwhile, our Win Transformer achieves consistently superior performance than Swin Transformer on multiple computer vision tasks, including image recognition, semantic segmentation,  and object detection.


\end{abstract}

\newpage

\section{Introduction}

Recently, the triumph achieved by Transformer~\citep{vaswani2017attention} in natural language processing, some works attempt to deploy Transformer in computer vision applications and built Vision Transformers~\citep{dosovitskiy2021image,touvron2021training,han2020survey}. Compared with convolutional neural networks~\citep{krizhevsky2012imagenet,he2016deep}, a Vision Transformer  has a larger receptive field, effectively capturing the long-range relations between pixels.  Meanwhile, different from convolution layer processing images by a static convolution kernel, the self-attention operations in Vision Transformer are dynamic, which effectively model the relations in the image adaptively. Nevertheless, the powerful representing capacity of the self-attention operations in Vision Transformers is accompanied by the potential efficiency issue. To be specific, the computational complexity of a self-attention operation is proportional to the square of the number of tokens. Thus,  only a limited number of tokens can be allowed in the self-attention operation to achieve satisfactory efficiency. The pioneering works, ViT~\citep{dosovitskiy2021image} and DeiT~\citep{touvron2021training}, considering the  efficiency, have to partition an image into a low-resolution feature map consisting of only $16\times 16$ patches/tokens. Nevertheless, the coarsely partitioned low-resolution feature map limits Vision Transformers' capability in visual understanding, especially for  dense prediction tasks such as  segmentation.

For effective dense prediction and acceptable efficiency, Pyramid Vision Transformer (PVT)~\citep{wang2021pyramid}, Pooling-based Vision Transformer (PiT)~\citep{heo2021rethinking} and Hierarchical Visual Transformer (HVT)~\citep{pan2021scalable} design a high-resolution feature map in the initial layer and reduces the resolution in deep layers with a progressive shrinking pyramid structure. In parallel to the pyramid structure, window-based local attention is another widely used manner for improving  efficiency. Swin Transformer~\citep{liu2021Swin}  crops an image into non-overlapping windows, and the self-attention is exploited within each window. In spite of improving  efficiency  considerably, the attention within the  window  constraints that a patch can only  perceive the patches within the same window. The limited perceptive field impedes the communications between patches from different  windows and might deteriorate  understanding capability for long-range relations. To enable  cross-window communication,  Swin Transformer devises a shifted  window partitioning mechanism. It adopts different manners for window partitioning in two consecutive layers, and the receptive field of each patch is enlarged. Similarly, Twins~\citep{chu2021twins} and Shuffle Transformer~\citep{huang2021shuffle} also  adopt the window-based local attention for high efficiency and devote a lot of effort to bringing back the global receptive field for high effectiveness. Specifically,  Twins~\citep{chu2021twins} develops  global sub-sampled attention, which is with global receptive field and complements window-based local attention. In a similar manner, Shuffle Transformer develops a shuffled local-window attention module, which  is complementary to the vanilla window-based local attention.  

\vspace{0.1in}

In this work, we check the necessity of the sophisticated operation, shifted window partitioning,  for bringing back the global receptive field proposed  in Swin~\citep{liu2021Swin}. 
To differentiate from the shifted window-based local attention, we term the vanilla window-based local attention without shifting as plain window-based local attention.
Surprisingly, we discover that, after pairing a plain window-based local attention layer  with a simple depthwise convolution, it has consistently better  performance than Swin Transformer in image recognition, semantic segmentation, and object detection. Meanwhile, with the existence of  depthwise convolutions,  the shifted window partitioning in Swin can not  bring additional performance improvement. Thus, we degenerate Swin to a plain Window-attention Transformer (Win), which is  conceptually simpler than Swin, Twins, and Shuffle Transformer. Specifically, we have removed quite a few lines of code in Swin Transformer. Meanwhile, our Win Transformer achieves  competitive performance in image recognition, object detection, and semantic segmentation. Compared with previous Vision Transformers, the proposed Win Transformer does contain any sophisticated operations. On the contrary, it removes some redundant operations. Taking the simplicity, the efficiency, and the effectiveness into consideration, our Win Transformer is a good choice for  real applications.



%


\section{Related Work}

\textbf{Self-attention mechanism in CNN.} In the past decade, the convolutional neural network (CNN)~\citep{lecun1995convolutional,krizhevsky2012imagenet,he2016deep} has  been \emph{de facto} backbone for vision understanding. The local receptive field of the convolution layer limits its capability of capturing the long-range dependencies. Thus, there are many works attempting to improve the performance of CNN through introducing the self-attention operation. Non-local neural network~\citep{wang2018non} is one of the earliest works to plug self-attention modules in CNNs for video understanding. Since the long-range dependencies are important for video recognition,  self-attention is considerably helpful in this scenario. \citet{cao2019gcnet,yin2020disentangled} further investigate the non-local neural network in depth and apply it in various image understanding tasks such as semantic segmentation and object detection. Attention Augmented Convolutional Network~\citep{bello2019attention} builds a hybrid architecture  by replacing a few convolution layers with self-attention layers.  It augments a convolution with the self-attention mechanism.  DETR~\citep{carion2020end}  builds an encoder-decoder Transformer architecture to post-process the feature map from a CNN for the object detection task.
Similarly, Relationnet++~\citep{chi2020relationnet++} builds an object detector through Transformer decoder. Deformable DETR~\citep{zhu2020deformable} improves the training efficiency and detection accuracy of DETR by an iterative bounding box refinement mechanism. SETR~\citep{zheng2021rethinking} adopts ViT~\citep{dosovitskiy2021image} as encoder and incorporates CNN decoders to inflate the feature resolution.  BoTNet~\citep{srinivas2021bottleneck} replaces the  convolutions with global self-attention in the last a few bottleneck blocks of a ResNet, achieving outstanding performance in image recognition. 

\vspace{0.2in}
\noindent \textbf{Global-attention Vision Transformer.} ViT~\citep{dosovitskiy2021image} builds the first pure Transformer architecture for computer vision tasks.  It tokenizes an image by cropping an image into $16\times 16$ patches/tokens and builds a pure Transformer-based architecture to process the tokens for image recognition. DeiT~\citep{touvron2021training} improves the data efficiency of ViT by introducing more advanced data augmentation and regularization methods.
Due to the self-attention operations' quadratic complexity  with respect to the number of tokens, ViT and DeiT can only adopt the low-resolution feature map, limiting their performance for dense prediction.  PVT~\citep{wang2021pyramid} improves the performance of Vision Transformers for dense prediction by  adopting a high-resolution feature map. To alleviate the high computational cost issue in a high-resolution feature map, PVT learns from the design of  the convolutional neural network and builds a hierarchical pyramid structure, progressively reducing the resolution to boost efficiency. PiT~\citep{heo2021rethinking}  and HVT~\citep{pan2021scalable} conduct pooling to   shrink the spatial dimension and inflates the channel dimension gradually for achieving high efficiency and effectiveness. Similarly, Multi-scale Vision Transformer~\citep{fan2021multiscale} hierarchically expands the channel capacity while reducing the spatial resolution.  Segformer~\citep{xie2021segformer} adopts the hierarchically structured Transformer encoder which outputs multi-scale features for effective semantic segmentation. Tokens-to-Token ViT~\citep{yuan2021tokens}   recursively aggregates neighboring tokens into one token to model the local structure and progressively reduces the token length. Transformer iN Transformer (TNT)~\citep{han2021transformer} models the  attention inside these local patches to  boost the performance further. CaiT~\citep{touvron2021going} and DeepViT~\citep{zhou2021deepvit} investigate in  deeper Transformer for achieving higher recognition accuracy.  Convolutional vision Transformer (CvT)~\citep{wu2021cvt} introduces convolutions to capture local spatial context and reduce semantic ambiguity in the attention mechanism. LeViT~\citep{graham2021levit}, Early Convolutions~\citep{xiao2021early}, Uniformer~\citep{li2022uniformer}  and LIT~\citep{pan2022less} adopt  convolutions/MLPs in early stages to improve Vision Transformers' efficiency and effectiveness. 

\vspace{0.1in}
\noindent \textbf{Local-attention Vision Transformer.} To further improve efficiency, some recent works~\citep{liu2021Swin,chu2021twins,huang2021shuffle} exploit window-based local attention. They only exploit the self-attention operation within each local window individually, and the computational burden is alleviated. Nevertheless,  window-based local attention impedes the communications between patches located in different windows and thus discards the long-range dependencies. To overcome the drawbacks inherited in  window-based local attention, Swin Transformer~\citep{liu2021Swin}, Twins~\citep{chu2021twins}, and Shuffle Transformer~\citep{huang2021shuffle}  adopt a dual-block architecture. That is, for consecutive two blocks, the first block adopts the  window-based local attention, and the second one attempts to achieve the cross-window communications. To be specific, Swin Transformer~\citep{liu2021Swin}
adopts the shifted window partition in the second block. 
Twins~\citep{chu2021twins} devises global sub-sampled attention in the second block. 
The global sub-sampled attention block  down-samples the global feature map into a low-resolution feature map to achieve a global receptive field in a lightweight manner.   Shuffle Transformer~\citep{huang2021shuffle} designs a  spatial shuffle operation into the second window-based self-attention block for building connections among windows.  Different from the dual-block architecture  achieving the large receptive field through a cascade of two complementary blocks, recent Vision Transformers~\citep{yang2021focal,dong2021cswin} put efforts on achieving the satisfactory receptive field within every single block in a parallel manner. Specifically, in Focal Transformer~\citep{yang2021focal}, a patch attends its closest surroundings in the same local window.
In the meanwhile,  it also sparsely sub-samples  patches far from the patch as  complementary information. CSWin~\citep{dong2021cswin}  computes
self-attention in the horizontal and vertical stripes in parallel and forms a cross-shape window. 
DW-S Conv~\citep{han2021demystifying} attempts to replace the self-attention operations in the local Vision Transformer with cheaper dynamic depthwise convolution, achieving comparable performance. BOAT~\citep{yu2022boat} exploits the feature-space local attention to boost the local-attention vision Transformer. MVT~\citep{chen2021mvt} utilizes the locality in projected view and proposed a local-global structure for effective and efficient 3D object recognition.

\vspace{0.2in}
\noindent \textbf{Applications of Transformer at Baidu.} At Baidu, transformer and related techniques have been widely applied for (e.g.,) advertising, especially image/video advertising, including multi-modal/cross-modal retrieval for advertising~\citep{yu2020combo,yu2021heterogeneous,yu2021tira,yu2022boost}. With the widespread use of transformer and other popular models in production, the issue of transformer-related model security has become an urgent matter. Therefore, recently there have been  major efforts at Baidu for developing techniques for AI model security~\citep{doan2021lira,doan2021backdoor,doan2022defending}.

\newpage

\section{Plain  Window-attention Vision Transformer}

\begin{figure}[h]

    \includegraphics[width=6.6in]{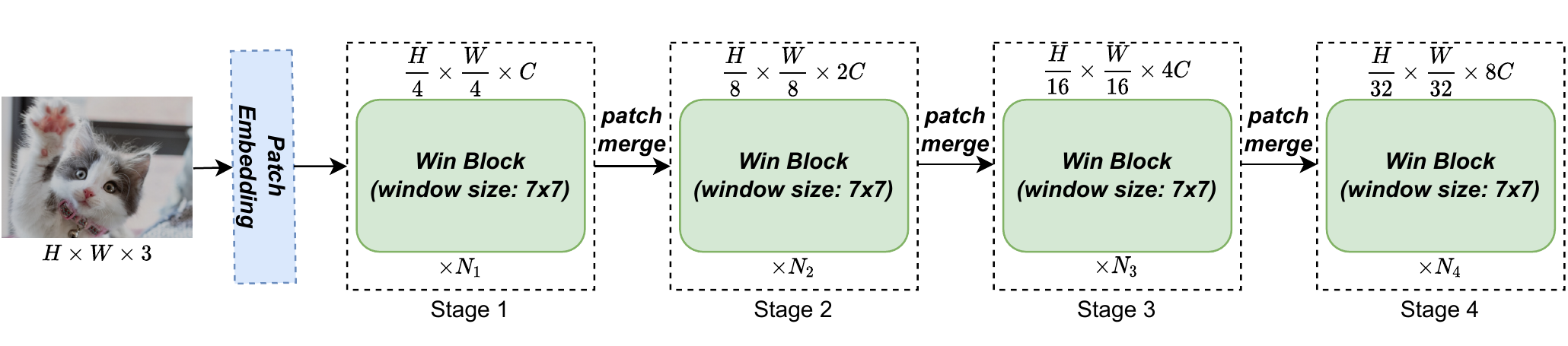}

\vspace{-0.1in}

    \caption{The architecture of the proposed plain Window-attention (Win) Transformer. It adopts a hierarchical pyramid structure with four stages.}
    \label{fig:arch}\vspace{0.1in}
\end{figure}

\noindent \textbf{Overall architecture.} Following the previous works, Swin Transformer~\citep{liu2021Swin} and Shuffle Transformer~\citep{huang2021shuffle},  the proposed plain Window-attention (Win) Transformer also adopts a hierarchical pyramid structure consisting of four stages as visualized in Figure~\ref{fig:arch}.

Given an image of $H\times W \times C$ size, a patch embedding layer generates the initial feature map of $\frac{H}{4} \times \frac{W}{4} \times C$ size. Between consecutive two stages, there is a patch merging layer to reduce to half the width/height of the feature map and increases the channel dimension by $2$. To be specific, in the $i$-th stage, the feature map is of $\frac{W}{2^{i+1}} \times \frac{H}{2^{i+1}} \times 2^{i-1}C $ size. Each stage consists of a stack of the proposed Window-attention (Win)  Block.

\vspace{0.1in} \noindent \textbf{Patch embedding and merging.} The path embedding module  converts a raw image into a feature map for further processing.  Following Swin~\citep{liu2021Swin}, in the patch embedding module, it crops an image into $4\times 4$ non-overlapped patches. A fully-connected layer projects each $4\times 4$  patches with $3$ channels into a $C$-dimension vector.  Between  two consecutive stages, we plug in a patch merging module to reduce the spatial resolution and increase the channel number.
Given a $W\times H \times C$ feature map in the input,  the output  of the patch merging is of $\frac{H}{2}\times \frac{W}{2}\times D$ size.  Similar to Swin~\citep{liu2021Swin}, the patch merging module   concatenates the
features of each group of $2\times 2$ $D$-dimension patch features into a $4D$-dimension feature vector and further projects it into a $2D$-dimension vector. 

\begin{figure}
    \centering
    \includegraphics[width=6in]{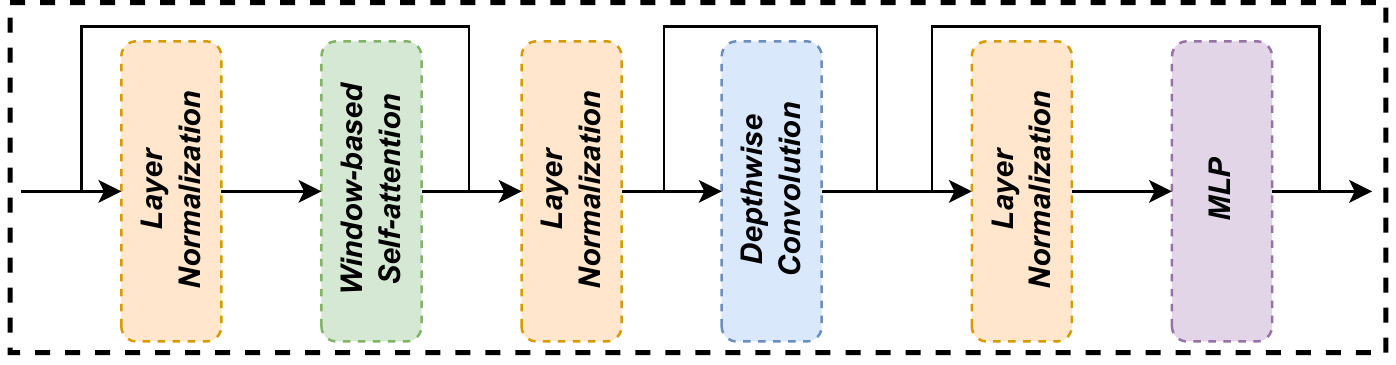}

\vspace{-0.1in}

    \caption{The Window-attention (Win) block consists of a window-based self-attention module, a depthwise convolution layer, an MLP module, and several layer-normalization layers. }
    \label{fig:str}
\end{figure}

\vspace{0.1in}
\noindent \textbf{Window-attention (Win) block.} As shown in Figure~\ref{fig:str}, a Win block  is built upon several layer-normalization (LN) layers~\citep{ba2016layer},  a window-based self-attention (WSA) module, an MLP module, a depthwise convolution (DConv) layer, and several skip connections. Given an input feature map $\mathcal{X} \in \mathbb{R}^{h\times w \times c}$, it obtains the output $\mathcal{Y}$ by conducting the below operations sequentially:
\begin{equation}
\begin{split}
    \mathcal{Y} &\gets \mathcal{X} + \mathrm{WSA}(\mathrm{LN}(\mathcal{X})),  \\
    \mathcal{Y} &\gets \mathrm{LN}(\mathcal{Y}) + \mathrm{DConv}(\mathrm{LN}(\mathcal{Y})),  \\
     \mathcal{Y} &\gets \mathcal{Y} + \mathrm{MLP}(\mathrm{LN}(\mathcal{Y})).  \\
 \end{split}   
\end{equation}
The WSA module adopts the same setting as Swin~\citep{liu2021Swin} which uniformly partitions each feature map into non-overlapping  $7\times 7$ windows. The multi-head self-attention operation is conducted within each $7\times 7$ local window. In the meanwhile, we add a position encoding operation in the window-based self-attention layer. We have attempted two types of position encoding,  including the relative positional encoding (RPE)  used in Swin~\citep{liu2021Swin}  and locally-enhanced positional encoding (LEPE) in CSWin~\citep{dong2021cswin}. The experiments show that  RPE can perform slightly better  than LEPE in image recognition on the ImageNet1K dataset. In contrast, for the downstream tasks such as semantic segmentation and object detection, LEPE performs slightly better than RPE.   By default, we adopt LEPE on all experiments.  To achieve the cross-window communications, we simply add a depth-wise convolution with a layer normalization and a residual connection after the window-based self-attention layer. The kernel size of the depthwise convolution is $7\times 7$.  The MLP module consists of two fully-connected layers with a GELU activation layer between them. The first fully-connected layer inflates the channel number of each patch from $C$ to $rC$, and the second MLP layer shrinks the channel number from $rC$ back to $C$, where $r>1$ is the inflation ratio. By default, we set $r=4$ in all settings, following Swin Transformers.

\begin{figure}[h]
\mbox{
     \subfigure[late residual]{
     \includegraphics[scale=0.9]{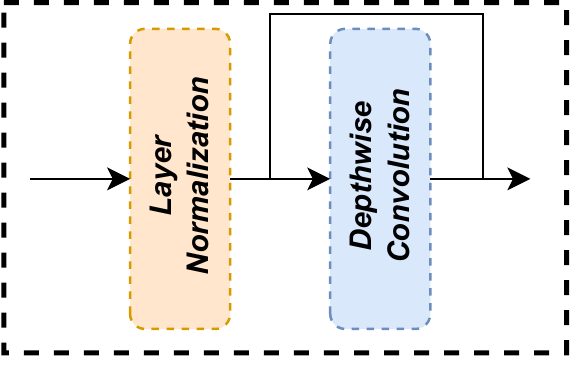}}
     \subfigure[early residual]{
     \includegraphics[scale=0.9]{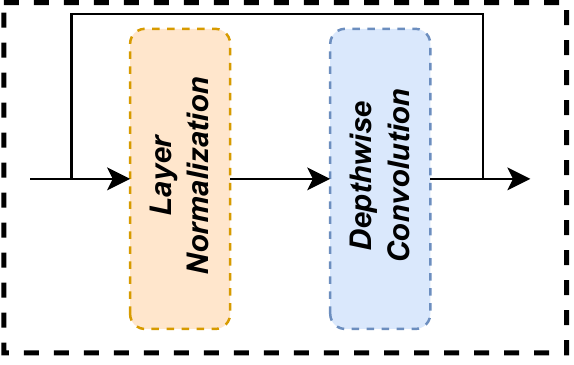}
     }
     \subfigure[w/o residual]{
     \includegraphics[scale=0.9]{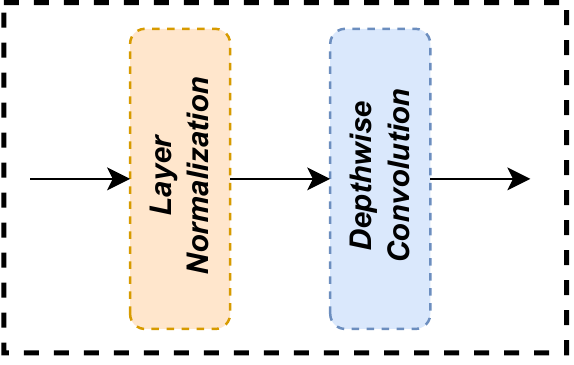}
     }
}

\vspace{-0.1in}

     \caption{Three different approaches for incorporating the depthwise convolution.}
      \label{fig:dep}
 \end{figure}

\vspace{0.1in}
\noindent \textbf{Discussion.} Our Win Block can be regarded as a Swin Block without shifting. The only add-on module is a depthwise convolution, and our Win block does not conduct any novel and sophisticated operations. We attempt three types of ways to incorporate the depthwise convolution as visualized in Figure~\ref{fig:dep}. 

Our experiments show the late-residual manner in Figure~\ref{fig:dep} (a)  achieves the best performance. Actually, the manner of integrating depthwise convolution has a considerable influence on the model performance.
Through extensive experiments,  we discover that the simple depthwise convolution is sufficient for   effective cross-window communications. With the existence of  depthwise convolutions,  sophisticated operations such as the shifted window partitioning in Swin Transformer can not bring further performance improvement. In fact,  depthwise convolutions have been exploited in Shuffle Transformer to boost the model's performance. Nevertheless, Shuffle Transformer  uses  depthwise convolutions to complement the shuffled window attention for cross-window communication. In contrast, our Win Transformer only uses the plain local window attention and solely relies on the depthwise convolution for cross-window communications.  Moreover, Shuffle Transformer adopts the early-residual manner as shown in Figure~\ref{fig:dep} (b). Our experiments show that  the early-residual manner can not achieve an excellent performance as the late-residual manner in our Win Transformer.


\vspace{0.1in}
\noindent \textbf{Complexity analysis.} Due to using an additional depthwise convolution,  our Win block takes slightly fewer parameters and more FLOPs than the Swin block. But the depthwise convolution is a very lightweight operation, which takes only a small number of parameters and FLOPs.
We denote the channel number by $c$, the spatial size of the depthwise convolution kernel by $k\times k$, and the spatial size of the feature map size by $H\times W$. In this case, the depthwise convolution only takes $k^2c$  parameters and $WHk^2c$ FLOPs, which are marginal with respect to the total number of parameters and FLOPs of a Vision Transformer block.

\section{Experiments}

Following Swin~\citep{liu2021Swin}, we built three types of backbones, including the tiny model (Win-T), the small model (Win-S), and  the base model
 (Win-B). We illustrate the detailed configurations of them in Table~\ref{tab:conf}. They adopt  similar architectures as their Swin counterparts. Compared with Swin, our Win takes  slightly more parameters and FLOPs. The marginal increases in the number of parameters and FLOPs are contributed to the lightweight depthwise convolution. The backbones are pre-trained on ImageNet-1K datasets without external data and are further fine-tuned on downstream tasks such as semantic segmentation and object detection.

\begin{table}[htpb!]
\centering
\caption{The configurations of Win-T, Win-S, and Win-B models.  $c$ denotes the channel number and $h$ denotes the head number in self-attention operation.\vspace{0.1in}}
\begin{tabular}{c|c|c|c|c}
\hline
Stage      &  Size &     Win-T & Win-S & Win-B \\ \hline
1          &  $56\times 56$           &  $\{c=96,h=3\} \times 2$    &  $\{c=96,h=3\} \times 2$       &   $\{c=128,h=4\} \times 2$     \\ 
2          &    $28\times 28$           &    $\{c=192,h=6\} \times 2$    &   $\{c=192,h=6\} \times 2$     &   $\{c=256,h=8\} \times 2$    \\ 
3          &    $14\times 14$           &     $\{c=384,h=12\} \times 6$    &   $\{c=384,h=12\} \times 18$    &   $\{c=512,h=16\} \times 18$     \\ 
4          &    $7\times 7$           &   $\{c=768,h=24\} \times 2$   &     $\{c=768,h=24\} \times 2$   &  $\{c=1024,h=32\} \times 2$      \\ \hline
\multicolumn{2}{c|}{Parameters}            & $29$ M     &  $50$ M       &  $88$ M    \\ \hline
\multicolumn{2}{c|}{FLOPs}           & $4.5$ G     &  $8.8$ G      &   $15.6$ G   \\ \hline
\end{tabular}
\label{tab:conf}
\end{table}

\subsection{Image classification}

\textbf{Settings.} We train the proposed Win Transformer on ImageNet1K dataset with $1.28$ million training images and 50 thousand images for validation from $1,000$ classes. Following DeiT~\citep{touvron2021training} and Swin~\citep{liu2021Swin},  we adopt AdamW~\citep{loshchilov2019decoupled} optimizer and train our models for $300$ epochs using a cosine decay learning rate scheduler. We also  conduct linear warm-up for the first $20$ epochs. We set the batch size as $1024$, the initial learning rate as $0.001$, and the weight decay as $0.05$. Following Swin,  we adopt several data augmentation and regularization methods including   RandAugment~\citep{cubuk2020randaugment}, Mixup~\citep{zhang2018mixup}, Cutmix~\citep{yun2019cutmix}, random erasing~\citep{zhong2020random} and stochastic depth~\citep{huang2016deep}. We do not use repeated augmentation~\citep{hoffer2020augment} and Exponential Moving Average (EMA)~\citep{polyak1992acceleration}, either.
We set the stochastic depth with a ratio of $0.1$, $0.3$, $0.5$ for  Win-T, Win-S, and Win-B, respectively.

\begin{table}[!t]
\centering
\caption{Comparisons with convolutional neural networks and other Vision Transformers on ImageNet-1K image classification. \vspace{0.1in}}
\begin{tabular}{c|c|c|c|c}
\toprule[1pt]
Method  & Input Size&  Params. (M)  & FLOPs (G)  & Top-1 Acc. ($\%$) \\ 
\midrule
ReGNetY-4G~\citep{radosavovic2020designing} & 224 & 21 & 4.0 & 80.0\\ 
ConvNeXt-T~\citep{liu2022convnet} & 224 & 29  & 4.5  & 82.1\\
DeiT-S~\citep{touvron2021training} & 224 & 22  & 4.6 & 79.8   \\
PVT-S~\citep{wang2021pyramid} & 224 & 25 & 3.8 & 79.8 \\
PiT-S~\citep{heo2021rethinking} & 224 & 24 & 2.9 & 80.9 \\
T2T-14~\citep{yuan2021tokens}  & 224 & 22 & 5.2 & 81.5 \\
TNT-S~\citep{han2021transformer}  & 224 & 24 & 5.2 & 81.3 \\
Swin-T~\citep{liu2021Swin} & 224 & 29  & 4.5  & 81.3   \\
Shuffle-T~\citep{huang2021shuffle} & 224 & 29 & 4.6 & 82.5 \\ 
NesT-T~\citep{zhang2022nested} & 224 & 17 & 5.8 & 81.5 \\
Focal-T~\citep{yang2021focal} & 224 &  29 & 4.9   &  {82.2} \\
CrossFormer-S~\citep{wang2022crossformer} & 224 &  31 & 4.9   &  {82.5} \\
CSWin-T~\citep{dong2021cswin}  & 224 & 23  & 4.3  & \textbf{82.7}   \\
Win-T (ours)  & 224 & 29  & 4.6 & {82.3}   \\
\midrule
ReGNetY-8G~\citep{radosavovic2020designing} & 224 & 39 & 8.0 & 81.7 \\
ConvNeXt-T~\citep{liu2022convnet} & 224 & 50 & 8.7 & 83.1\\
PVT-L~\citep{wang2021pyramid} & 224 & 61 & 9.8 & 81.7 \\
T2T-19~\citep{yuan2021tokens}  & 224 & 39 & 8.9 & 81.9 \\
MViT-B~\citep{fan2021multiscale} & 224 & 37 & 7.8 & 83.0 \\
Swin-S~\citep{liu2021Swin}  & 224 & 50  & 8.7  & 83.0   \\
Twins-B~\citep{chu2021twins} & 224 & 56 & 8.3 & 83.2 \\
Shuffle-S~\citep{huang2021shuffle} & 224 & 50 & 8.9 & 83.5 \\ 
NesT-S~\citep{zhang2022nested} & 224 & 38 & 10.4 & 83.3 \\
Focal-S~\citep{yang2021focal} & 224 &  51 & 9.1   &  {83.5} \\
CrossFormer-B~\citep{wang2022crossformer} & 224 &  52 & 9.2   &  {83.4} \\
CSWin-S~\citep{dong2021cswin}   & 224 & 35  & 6.9 & 83.6   \\
Win-S (ours)  & 224 &   50  & 8.9 & \textbf{83.7}   \\
\midrule
ConvNeXt-B~\citep{liu2022convnet} & 224 & 89 & 15.4 & 83.8\\
DeiT-B~\citep{touvron2021training}  & 384 & 86  & 55.4  & 83.1   \\
DeiT-B~\citep{touvron2021training}  & 224 & 86  & 17.5  & 81.8   \\
PiT-B~\citep{heo2021rethinking} & 224 & 74 & 12.5 & 82.0 \\
Swin-B~\citep{liu2021Swin} & 224 & 88  & 15.4 & 83.5   \\
Twins-L~\citep{chu2021twins} & 224 & 99 & 14.8 & 83.7 \\
Shuffle-B~\citep{huang2021shuffle} & 224 & 88 & 15.6 & 84.0 \\ 
NesT-B~\citep{zhang2022nested} & 224 & 68 & 17.9 & 83.8 \\
Focal-B~\citep{yang2021focal} & 224 &  90 & 16.0   &  {83.8}    \\
CrossFormer-L~\citep{wang2022crossformer} & 224 &  92 & 16.1   &  {84.0} \\
CSWin-B~\citep{dong2021cswin}  & 224 & 78  & 15.0  & \textbf{84.2} \\
Win-B (ours)  & 224 &   88  & 15.6  & 83.5   \\

\bottomrule[1pt]
\end{tabular}
\label{tab:class}\vspace{0.1in}
\end{table}

\vspace{0.1in}
\noindent \textbf{Comparisons with existing methods.} To demonstrate the effectiveness of the proposed Win in the image recognition task, we compare it with a representative convolution neural network, ReGNet~\citep{radosavovic2020designing}, and other Vision Transformers, including 
DeiT~\citep{touvron2021training}, PVT~\citep{wang2021pyramid}, PiT~\citep{heo2021rethinking}, T2T~\citep{yuan2021tokens},  Swin Transformer~\citep{liu2021Swin}, Shuffle Transformer~\citep{huang2021shuffle}, NesT~\citep{zhang2022nested}, Focal Transformer~\citep{yang2021focal}, CrossFormer~\citep{wang2022crossformer}, and CSWin Transformer~\citep{dong2021cswin}. As shown in Table~\ref{tab:class}, using comparable number of parameters and FLOPs, our Win-T, Win-S, and Win-B achieve competitive performance compared with their counterparts. Especially, our  
Win-S achieves the best performance among all small-scale models. Note that our Win Transformers do  not contain any sophisticated operations and are simply built upon the plain window-based self-attention layers, depthwise convolution layers, and MLP layers. In contrast, the compared counterparts devise several sophisticated  operations, which are more complicated. It is considerably more difficult to implement them compared with ours.
Note that, as shown in Table~\ref{tab:class}, our   medium-scale model, Win-B, performs a little bit worse than the smaller counterpart, Win-S. The worse performance might be due to over-fitting.

\vspace{0.1in}
\noindent 
\textbf{The manner of integrating depthwise convolution. } We evaluate the performance of different 
manners for integrating depthwise convolution. We compare three  manners as visualized in Figure~\ref{fig:dep}, including late residual connection  in Figure~\ref{fig:dep} (a),  early residual connection  in Figure~\ref{fig:dep} (b),  and w/o residual connection in Figure~\ref{fig:dep} (c). As shown in Table~\ref{tab:inter},  the late residual connection achieves the best performance. Besides, after removing the residual connections, the performance  considerably deteriorates. In the meanwhile, we observe from the table that, the advantage of the last residual connection over the early residual connection is more significant as the model capacity increases.

\begin{table}[htpb!]
\centering
\caption{The performance of different manners for integrating depthwise convolution. We report the top-1 accuracy of image recognition on ImageNet-1K dataset.\vspace{0.1in}}
\begin{tabular}{c|c|c|c}
\hline
Model    &  late residual &     early residual & w/o residual \\ \hline
Win-T     &  $82.3$    &   $82.2$            &  $82.1$ \\ 
Win-S    &  $83.7$    &   $83.3$     &  $83.0$    \\ 
Win-B       & $83.5$          &    $83.1$     &  $83.0$     \\ \hline
\end{tabular}
\label{tab:inter}
\end{table}

\vspace{0.1in}
\noindent \textbf{Depthwise convolution and shifted window partitioning.} We use depthwise convolution for achieving  cross-window communication. In contrast, Swin adopts the shifted window partitioning for communications between patches from different local windows. Here, we evaluate  effectiveness of 
 depthwise convolution in our Win and  the shifted window partitioning in Swin. 
 
 \begin{table}[htpb!]
\centering
\caption{The influence of depthwise convolution and shifted window partitioning. We report the top-1 accuracy of image recognition on ImageNet-1K dataset.\vspace{0.1in} }
\begin{tabular}{c|cccc}
\hline
Depthwise Conv  &  &   &   \checkmark  &  \checkmark\\ 
Shifted Window    &      &  \checkmark             & & \checkmark \\ \hline
Win-T & $81.1$ & $81.3$ &  $82.3$& $82.2$\\
Win-S & $82.3$ & $83.0$ & $83.7$ & $83.7$ \\ \hline
\end{tabular}
\label{tab:conv}
\end{table}
 
 As shown in Table~\ref{tab:conv}, by removing both depthwise convolution and shifted window partitioning, Win-Tiny as well as Win-Small cannot achieve a satisfactory image recognition performance. The low recognition accuracy is due to a lack of cross-window communication.  After adopting the depthwise convolution or shifted window partitioning to achieve cross-window communication, the image recognition performance  gets improved consistently for both Win-Tiny and Win-Small models.  Table~\ref{tab:conv} illustrates that the depthwise convolution is more effective than the shifted window partitioning, and it achieves a considerably higher image recognition accuracy. In the meanwhile, with the existence of the depthwise convolution, the shifted window partition can not bring further performance improvement. Therefore, we do not adopt  the shifted window partitioning in our Win and just adopt the plain local window partitioning.

\vspace{0.1in} 
\noindent \textbf{The kernel size of depthwise convolution.} We evaluate the influence of the kernel size of the depthwise convolution. We vary the kernel size among $\{3\times 3, 5\times 5, 7\times 7\}$. Our experiments in Table~\ref{tab:size} shows that, the Win-Tiny model using a  depthwise convolution with only $3\times 3$ spatial size has  achieved excellent performance on image recognition. When the spatial size increases from $3\times 3$ to $7\times 7$, the performance slightly improves. In the meanwhile, as the spatial size of the depthwise increases, the number of parameters and  FLOPs also increase.  Since the increase in the number of parameters and  FLOPs is marginal, we set the large kernel with $7\times 7$ spatial size by default.

 \begin{table}[htpb!]
\centering
\caption{The influence of the depthwise convolution kernel size. We report the performance of Win-Tiny on ImageNet-1K dataset.\vspace{0.1in} }
\begin{tabular}{c|c|c|c}
\hline
kernel size  & Params. (M)  &   FLOPs (G) &  Top-1 Acc. ($\%$)  \\ \hline
$3\times 3$    & $28.33$     & $4.51$ &  $82.2$ \\ 
$5\times 5$ &$28.36$  &$4.52$ & $82.2$ \\
$7\times 7$ & $28.50$ & $4.56$& $82.3$ \\ \hline
\end{tabular}
\label{tab:size}
\end{table}

\vspace{0.1in}
\noindent \textbf{The influence of the positional encoding.} We attempt different types of positional encoding, including the relative positional encoding (RPE) used in Swin and locally-enhanced positional encoding (LEPE) in CSWin. In fact, RPE is equivalent to a $7\times 7$ depthwise convolution with zero padding.  In contrast, LEPE is implemented by a $3\times 3$ depthwise convolution.  

 \begin{table}[htpb!]
\centering
\caption{The influence of the positional encoding. We report the top-1 classification accuracy of Win-Tiny and Win-Small on ImageNet-1K dataset. \vspace{0.1in} }
\begin{tabular}{c|c|c|c}
\hline
Model & RPE & LEPE &   w/o PE   \\ \hline
Win-T    & $82.33$  & $82.30$ & $82.22$ \\ 
Win-S & $83.75$ & $83.67$ & $83.55$ \\\hline
\end{tabular}
\label{tab:pos}
\end{table}

As shown in Table~\ref{tab:pos}, RPE achieves slightly higher recognition accuracy than LEPE for both Win-Tiny and Win-Small models. The higher accuracy might be attributed to the larger receptive field of RPE.  In the meanwhile, we also testify the performance  of our Win without position coding (w/o PE). Table~\ref{tab:pos} illustrates that, after removing the positional encoding, the performance of the proposed Win  becomes slightly worse, but its accuracy is still satisfactory. This is might be due to the fact that the $7\times 7$ depthwise convolution we plug after the window-based self-attention module originally developed for achieving cross-window communications simultaneously has the capability of encoding the positional information. Despite that  RPE achieves a slightly better performance than LEPE in ImageNet1K image recognition, we set LEPE as the default positional encoding approach since it performs better in the downstream tasks.

\subsection{Semantic segmentation}
  
\textbf{Dataset.}  We evaluate the performance of the proposed Win in semantic segmentation on a widely used public benchmark, ADE20K~\citep{zhou2017scene}.   It consists of $20,210$  images for training, $2,000$  images for validation, and $3,352$ images for testing.   Following the exiting works, we use the training images to train our models and report the performance on the validation images.  In testing, we report the experimental results of models using the evaluation metric, mIoU, under  single-scale (SS) and multi-scale (MS) settings.  In the SS setting, the scale of testing images is the same as that of the training images.  
In contrast, the MS setting adopts scales that are $[0.5, 0.75, 1.0, 1.25, 1.5, 1.75] \times $ of that for training.

\vspace{0.1in}
\noindent \textbf{Setting.} We load the  models pre-trained on ImageNet-1K dataset and fine-tune them on the semantic segmentation. To compare with the existing state-of-the-art Vision Transformers in a fair manner,  we also adopt UpperNet~\citep{xiao2018unified} as the framework for segmentation. Following Swin, in training, we employ the AdamW~\citep{loshchilov2019decoupled} optimizer. The initial learning rate is set as $6 \times 10^{-5}$.  The weight decay is $0.01$. We adopt  a scheduler that uses linear learning rate decay. We use a linear warmup with $1,500$ iterations. We set the batch size as $16$ equally divided into $8$ GPU cards. 
All the models are trained with input size $512 \times 512$.
The whole training process lasts for $160$ thousand iterations.
In training,  we adopt multiple data augmentation methods, including random horizontal flipping, random re-scaling within ratio range $[0.5, 2.0]$, and random photometric distortion. We set the stochastic depth with a ratio of $0.2$ for all Win Transformer models.

 \begin{table}[!t]
\centering
\caption{Comparisons with the state-of-the-art convolutional neural networks and  Vision Transformers on ADE20K semantic segmentation. FLOPs is measured on the input image of $1024 \times  1024$ resolution. SS denotes the single-scale setting and MS denotes the multi-scale setting. \vspace{0.1in}}
\begin{tabular}{c|c|c|c|c}
\toprule[1pt]
Method  &  Params. (M)  & FLOPs (G)  & SS mIoU ($\%$)  & MS mIoU ($\%$) \\ 
\midrule
ConvNeXt-T~\citep{liu2022convnet} & 60 & 939 & - & 46.7 \\
TwinsP-S~\citep{chu2021twins} & 55 & 919 & 46.2 & 47.5 \\
Twins-S~\citep{chu2021twins} & 54 & 901 & 46.2 & 47.1\\
CSWin-T~\citep{dong2021cswin}  &  60 & 959  &  {49.3} & 50.4 \\
Shuffle-T~\citep{huang2021shuffle}  &  60 & 949   &  {46.6} & 47.6 \\
Focal-T~\citep{yang2021focal} & 62 & 998 & {45.8} & 47.0\\
Swin-T~\citep{dong2021cswin}   & 60  & 945  & {44.5} & 45.8   \\
Win-T (ours)   & 60  & 949 & {45.7} & 47.2  \\
\midrule
ConvNeXt-S~\citep{liu2022convnet} & 82 & 1027 & - & 49.6 \\
TwinsP-B~\citep{chu2021twins} & 74 & 977 & 47.1 & 48.4 \\
Twins-B~\citep{chu2021twins} & 89 & 1020 & 47.7  & 48.9\\
CSWin-S~\citep{dong2021cswin}  &  65 & 1027  &  {50.4} & 50.8 \\
Shuffle-S~\citep{huang2021shuffle}  &  81 & 1044   &  {48.4} & 49.6 \\
Focal-S~\citep{yang2021focal} & 85 & 1130  & {48.0} & 50.0  \\
Swin-S~\citep{dong2021cswin}   & 81  & 1038  & {47.6} & 49.4   \\
Win-S (ours)   & 81  & 1044 & {48.4} & 49.8  \\
\midrule
ConvNeXt-B~\citep{liu2022convnet} & 122 & 1170 & - & 49.9 \\
TwinsP-L~\citep{chu2021twins} & 92 & 1041 & 48.6 & 49.8 \\
Twins-L~\citep{chu2021twins} & 133 & 1164 & 48.8 & 50.2\\
CSWin-B~\citep{dong2021cswin}  &  109 & 1222  &  {51.1} & 51.7 \\
Shuffle-B~\citep{huang2021shuffle}  & 121   & 1196    &  {49.0} & 50.5 \\
Focal-B~\citep{yang2021focal} & 126 & 1354 & {49.0} & 50.5 \\
Swin-B~\citep{dong2021cswin}   & 121  & 1188  & {48.1} & 49.7   \\
Win-B (ours)   & 121  & 1196 & {49.2} & 50.7  \\

\bottomrule[1pt]
\end{tabular}
\label{tab:seg}\vspace{0.3in}
\end{table}

\vspace{0.1in}
\noindent \textbf{Comparisons with state-of-the-art models.} To demonstrate the excellence of the proposed Win Transformer in Semantic Segmentation, we compare
it with several state-of-the-art models including Twins~\citep{chu2021twins}, CSWin Transformer~\citep{dong2021cswin}, Shuffle Transformer~\citep{huang2021shuffle}, Focal Transformer~\citep{yang2021focal}, and Swin Transformer~\citep{liu2021Swin}. As shown in Table~\ref{tab:seg}, our Win Transforms have achieved competitive performance in ADE20K semantic segmentation compared with these state-of-the-art models. To be specific, our Win Transformer consistently outperforms Swin Transformer for tiny, small and base models on both SS and MS settings.  Meanwhile, on the SS setting, our Win-S outperforms Twins-B and Focal-S,  achieving the second-best performance among all the small-scale models. Moreover,  on the SS  and MS settings,  our  Win-B outperforms all compared models except the CSWin-B model and achieves the second-best performance among all the medium-scale models. Again, we emphasize that our Win Transformer is much simpler and easier for implementation than the compared Vision Transformers. The proposed Win Transformer does not need any sophisticated  operations but achieves excellent performance in semantic segmentation.

 \begin{table}[htpb!]
\centering
\caption{The influence of the positional encoding on ADE20K semantic segmentation. We report the mIoU on single-scale and multi-scale  settings.\vspace{0.1in}}
\begin{tabular}{c|cc|cc}
\hline
  & \multicolumn{2}{c|}{RPE}            & \multicolumn{2}{c}{LEPE}           \\ \cline{2-5}
  & \multicolumn{1}{c}{SS mIoU} & MS mIoU   & \multicolumn{1}{c}{SS mIoU} & MS mIoU \\ \hline
Win-T & \multicolumn{1}{c}{$45.3$}       &   $46.9$    & \multicolumn{1}{c}{$45.7$}       &  47.2     \\ 
Win-S & \multicolumn{1}{c}{$48.0$}       &  $49.6$     & \multicolumn{1}{c}{$48.4$}       &   49.8    \\ \hline
\end{tabular}
\label{tab:segpos}\vspace{0.2in}
\end{table}

\vspace{0.1in}
\noindent \textbf{The influence of the positional encoding.} We compare the performance of the relative positional encoding (RPE) and the locally-enhanced positional encoding (LEPE) on semantic segmentation. As shown in Table~\ref{tab:segpos}, using LEPE, Win-T as well as Win-S model  achieve a better performance in semantic segmentation. As we mentioned previously, compared with RPE, LEPE has a smaller receptive field, which might be helpful for semantic segmentation where the local context is critical.  By default, we adopt LEPE method.

\subsection{Object detection and instance segmentation}

\begin{table}[!t]
\centering
\caption{Comparisons with convolutional neural networks and Vision Transformers based on Mask-RCNN framework on MSCOCO-2017 object detection and instance segmentation.  FLOPs is evaluated on $1280 \times  800$
resolution.  AP$^b$  denotes the box-level average precision for evaluating the performance of object detection and   AP$^m$  denotes the mask-level average precision for   instance segmentation. \vspace{0.1in} }
\begin{tabular}{c|c|c|ccc|ccc}
\toprule[1pt]
Method  &  Params  & FLOPs  & AP$^b$  &   AP$^b_{50}$ &   AP$^b_{75}$ & AP$^m$  &   AP$^m_{50}$ &   AP$^m_{75}$ \\ 
\midrule
ResNet-50~\citep{he2016deep} & 44 M & 260 G & 41.0 & 61.7 & 44.9 & 37.1 & 58.4 & 40.1 \\
ConvNeXt-T~\citep{liu2022convnet} & 48 M & 262 G & 46.2 & 67.9 & 50.8 &  41.7 & 65.0 & 44.9 \\ 
PVT-S~\citep{wang2021pyramid} & 44 M & 256 G & 43.0 & 65.3 & 46.9 & 39.9 & 62.5 & 42.8 \\
ViL-S~\citep{zhang2021multi} & 45 M & 218 G & 47.1 & 68.7 & 51.5 & 42.7 & 65.9 & 46.2\\
TwinsP-S~\citep{chu2021twins} & 44 M & 245 G & 46.8 & 69.3 & 51.8 & 42.6 &66.3& 46.0 \\
Twins-S~\citep{chu2021twins} & 44 M & 228 G &46.8 & 69.2 & 51.2 & 42.6 & 66.3 & 45.8\\
CSWin-T~\citep{dong2021cswin}  & 42 M & 279 G   &  49.0 & 70.7 & 53.7 & 43.6 & 67.9 & 46.6 \\
Shuffle-T~\citep{huang2021shuffle}  & 48 M & 268 G& 46.8 & 68.9 & 51.5 & 42.3 & 66.0 & 45.6 \\
Focal-T~\citep{yang2021focal} & 49 M& 291 G & 47.2  & 69.4 & 51.9 & 42.7 & 66.5 & 45.9\\
Swin-T~\citep{dong2021cswin}   & 48 M& 264 G  & 46.0 & 68.2 & 50.2 & 41.6 & 65.1 & 44.8  \\
Win-T (ours)   & 48 M & 268 G  & 46.6 & 68.4 & 51.1 &  42.1 & 65.7 & 45.3\\
\midrule
ResNet-101~\citep{he2016deep} & 63 M & 336 G & 42.8 & 63.2 & 47.1 & 38.5 & 60.1 & 41.3\\
PVT-M~\citep{wang2021pyramid} & 64 M& 302 G& 44.2 & 66.0 & 48.2 & 40.5 & 63.1 & 43.5\\
ViL-M~\citep{zhang2021multi} & 60 M& 261 G  & 44.6 &  66.3 & 48.5 & 40.7 & 63.8 & 43.7\\
TwinsP-B~\citep{chu2021twins} & 64 M& 302 G& 47.9 & 70.1 & 52.5& 43.2 & 67.2 & 46.3\\
Twins-B~\citep{chu2021twins} & 76 M& 340 G & 48.0 & 69.5 & 52.7& 43.0& 66.8& 46.6\\
CSWin-S~\citep{dong2021cswin}  &  54 M & 342 G &  50.0 & 71.3 & 54.7 & 44.5 & 68.4 & 47.7\\
Shuffle-S~\citep{huang2021shuffle}  & 69 M & 359 G & 48.4 & 70.1 & 53.5 & 43.3 & 67.3 & 46.7 \\
Focal-S~\citep{yang2021focal} & 71 M& 401 G& 48.8 & 70.5 & 53.6 & 43.8 & 67.7 & 47.2 \\
Swin-S~\citep{dong2021cswin}   & 69 M& 354 G & 48.5 & 70.2 & 53.5 & 43.3 & 67.3 & 46.6   \\
Win-S (ours)   & 69 M & 359 G& {48.6} & 69.8 & 53.3 & 43.4 & 67.3  & 46.7  \\
\midrule
ResNeXt101-64~\citep{xie2017aggregated} & 101 M & 493 G & 44.4 & 64.9 & 48.8 & 39.7 & 61.9 & 42.6\\
PVT-L~\citep{wang2021pyramid} & 81 M & 364 G & 44.5 & 66.0 & 48.3 & 40.7 & 63.4  & 43.7\\
ViL-B~\citep{zhang2021multi} & 76 M & 365 G  & 45.7 & 67.2 & 49.9 & 41.3 & 64.4 & 44.5\\
CSWin-B~\citep{dong2021cswin}  &  97 M & 526 G  & 50.8 & 72.1 & 55.8 & 44.9 & 69.1 & 48.3 \\
Focal-B~\citep{yang2021focal} & 110 M  & 533 G &  49.0 & 70.1 & 53.6 & 43.7 & 67.6 & 47.0\\
Swin-B~\citep{dong2021cswin}   & 107 M & 496 G  & 48.5 & 69.8 & 53.2 & 43.4 & 66.8 & 46.9   \\
Win-B (ours)   & 108 M& 503 G& 49.1 & 70.2 & 53.8 & 43.8 & 67.8 & 47.1   \\
\bottomrule[1pt]
\end{tabular}
\label{tab:obj}
\end{table}

\textbf{Dataset.} We evaluate the object detection and instance segmentation on MS-COCO $2017$ public benchmark~\citep{lin2014microsoft}, consisting of  $118$ thousand training images, $5$ thousand validation images, and $20$ thousand test-dev images.  We pre-train the backbones on the ImageNet-1K dataset and then fine-tune the pre-trained model on the MS-COCO $2017$  training set.

\vspace{0.1in}
\noindent \textbf{Settings.} We adopt Mask-RCNN~\citep{he2017mask} as the object detection framework. Following Swin Transformer, we conduct  multi-scale training, which resizes the input image such that the shorter side is larger than $480$ and smaller than $800$ while the longer side is not larger than $1333$. We adopt a $3\times$ schedule with $36$ epochs in total and decay the learning rate to $0.1$ at the $27$th epoch and the $33$th epoch. We adopt  AdamW optimizer and set the initial learning rate as $0.0001$, weight decay as $0.05$, and batch size as $16$. We use $8$ GPU cards for training. We set the stochastic depth  ratio as $0.2$ for all Win Transformer models.

\newpage

\noindent \textbf{Comparisons with state-of-the-art methods.} To demonstrate the outstanding performance of the proposed Win Transformers on object detection and instance segmentation, we compare them with  convolutional neural networks including ResNet~\citep{he2016deep}, ResNeXt~\citep{xie2017aggregated} and ConvNeXt~\citep{liu2022convnet}. Meanwhile, we also compare them with  other Vision Transformers including PVT~\citep{wang2021pyramid}, ViL~\citep{zhang2021multi}, Twins~\citep{chu2021twins}, CSWin~\citep{dong2021cswin}, Shuffle Transformer~\citep{huang2021shuffle}, Focal Transformer~\citep{yang2021focal} and Swin Transformer~\citep{liu2021Swin}.  As shown in Table~\ref{tab:obj}, the proposed Win-T consistently  outperform the recent state-of-the-art convolutional neural network  (ConvNeXt-T) and Swin-T. In the meanwhile, our Win-S further outperforms  PVT-M, ViL-M, TwinsP-B, Shuffle-S, and Swin-B. Moreover, our Win-B achieves the second-best performance among all  medium-size Vision Transformers. Note that the proposed Win Transformer is much conceptually simpler than the compared Vision Transformers, and thus, its implementation is much easier. The competitive effectiveness, high efficiency, and  attractive simplicity make the proposed Win Transformer   a good choice in real applications.

\section{Conclusions} 

Vision Transformers suffer from the efficiency issue caused by  the quadratic computational complexity of self-attention operations with respect to the token number. 
To achieve a good trade-off between  effectiveness and efficiency,  some following works such as Swin Transformer adopt  window-based local attention mechanism. They  exploit the self-attention operations within each local window and attempt to achieve the cross-window communications through varying the window partitioning through several sophisticated operations, which brings difficulty in implementation. In this work, we attempt to achieve  cross-window communication through a  depthwise convolution  and check the necessity of these sophisticated operations.  Our experiments show that a simple depthwise convolution is sufficiently effective to achieve  cross-window communications.  With the existence of  depthwise convolution, the shifted window partitioning can not bring additional performance improvement. Thus, we remove the most innovative part in the Swin Transformer and degenerate the Swin Transformer to  the proposed Win Transformer. To demonstrate the excellence of the proposed Win Transformers, we conduct systematic experiments on multiple computer vision tasks including image recognition, semantic segmentation and object detection.  The experimental results  on multiple public benchmarks demonstrate the attractive effectiveness and efficiency of the proposed Win Transformers.

\bibliographystyle{plainnat}
\bibliography{refs_scholar}
\end{document}